%% file: main.tex
\documentclass[11pt,letterpaper]{article}

\usepackage[margin=1in]{geometry}
\usepackage[T1]{fontenc}
\usepackage[utf8]{inputenc}
\usepackage{lmodern}
\usepackage{microtype}
\usepackage{authblk}

\usepackage{eccvabbrv}

\usepackage{graphicx}
\usepackage{booktabs}

\usepackage[accsupp]{axessibility}  % Improves PDF readability for those with disabilities.

\usepackage{amsmath}
\usepackage{amssymb}
\usepackage{multirow}
\usepackage{xcolor}
\usepackage{enumitem}
\usepackage{siunitx}
\usepackage{tabularx}
\usepackage{array}
\usepackage{makecell}

\newcommand{\datasetname}{\textsc{OceanDepths}\xspace}
\usepackage{xspace}

\usepackage{pifont}

\definecolor{arxivblue}{RGB}{0,80,145}
\usepackage[breaklinks,colorlinks,allcolors=arxivblue]{hyperref}
\usepackage[capitalize,noabbrev]{cleveref}
\providecommand{\doi}[1]{\href{https://doi.org/#1}{\nolinkurl{https://doi.org/#1}}}

\usepackage{orcidlink}
\newcommand{\added}[1]{#1}

\begin{document}

% ---------------------------------------------------------------
\title{\datasetname: A Global Dataset of Paired Subsurface and Surface Ocean Observations} 

\author[1]{Simon Donike\orcidlink{0000-0002-4440-3835}}
\author[2]{Ruben Cartuyvels\orcidlink{0000-0003-1063-4659}}
\author[2]{Antonino Ian Ferola\orcidlink{0000-0003-3953-415X}}
\author[2]{Elisa Carli\orcidlink{0000-0001-8515-1606}}
\author[2]{Diego Fernandez Prieto\orcidlink{0000-0001-8984-6974}}
\author[2]{Marie-Helene Rio}
\affil[1]{Image Processing Laboratory, University of Valencia, Av. de Blasco Ibáñez 13, 46010, Spain\\\texttt{simon.donike@uv.es}}
\affil[2]{ESA-ESRIN, Via Galileo Galilei 1, 00044 Frascati RM, Italy}
\date{}

\maketitle

\input{sec/0_abstract}
\input{sec/1_intro}
\input{sec/2_related}

\input{sec/3_dataset}
\input{sec/4_evaluation}
% \input{sec/5_results}
\input{sec/6_conclusion}

% \section*{Acknowledgements}
% This dataset uses measurements made freely available by the \href{https://argo.ucsd.edu}{International Argo Program} and the \href{https://www.ocean-ops.org}{national programs} that contribute to it. The Argo Program is part of the Global Ocean Observing System. Additionally, it contains data collected from E.U. Copernicus Marine Service Information, licensed under the \href{https://marine.copernicus.eu/user-corner/service-commitments-and-licence}{Licence to Use Copernicus Marine Service Products}.

% ---- Bibliography ----
%
\bibliographystyle{splncs04}
\bibliography{main}

\newpage
\appendix

\input{sec/7_appendix}

%\newpage

%\input{sec/8_instructions}

% \section*{Acknowledgements}
% Please insert your acknowledgments here.

% % ---- Bibliography ----
% %
% % BibTeX users should specify bibliography style 'splncs04'.
% % References will then be sorted and formatted in the correct style.
% %
% \bibliographystyle{splncs04}
% \bibliography{main}
\end{document}

%% file: sec/0_abstract.tex
\begin{abstract}
Despite comprising over 70\% of its surface, the world's oceans are critically underobserved compared to the land surface or the atmosphere.
Understanding the global ocean requires jointly observing its surface and subsurface structure, yet no standardized, high-resolution dataset couples satellite surface fields to co-located \emph{in situ} depth profiles in an AI-ready format.
Existing resources either consist of model-reconstructed gridded products rather than observations, cover only a single variable or basin, or operate at resolutions too coarse for mesoscale dynamics.
We introduce \datasetname, the first open, global, regridded AI-ready dataset that pairs satellite-derived sea surface temperature (SST), sea surface salinity (SSS), and sea surface height (SSH) L4 products with co-located EN4 subsurface temperature and salinity profiles, complemented by matched GLORYS12 ocean reanalysis data to support comparisons or multi-stage learning.
The dataset spans 2000--2024 at \SI{0.1}{\degree}$\times$\SI{0.1}{\degree} spatial resolution and at weekly temporal resolution, covering the entire globe's sea surface and with over 9.5 million paired profiles interpolated to 50 standardized depth levels. 
We provide a configurable system to split the globe in equally sized spatial patches.
The 4D multivariate structure, high resolution, long temporal extent, and extreme sparsity of subsurface observations (${\sim}$0.01\% per depth level) make \datasetname a challenging testbed for novel AI methods.
We demonstrate subsurface state reconstruction as an example task with simple baseline models, but also envision \datasetname to support the development of observation-based forecast methods and other related tasks.
\added{Available at: \url{https://huggingface.co/datasets/ESA-philab/OceanDepths}.}
% lead to progress in oceanography.
% , as well as a valuable resource for the development of observation-based oceanography methods.
% well beyond the sparsity regimes explored by current approaches.
% We demonstrate two representative tasks (subsurface reconstruction and sea surface temperature forecasting), provide a review of related datasets and methods, and establish baselines for both tasks.
% \datasetname gives the ocean and computer vision communities a single, open benchmark for research spanning vertical reconstruction, spatiotemporal forecasting, sparse-data learning, and the study of global ocean dynamics.
\end{abstract}

%% file: sec/1_intro.tex
\section{Introduction}
\label{sec:intro}

The global ocean plays a central role in regulating Earth's climate, absorbing over 90\% of the excess heat trapped by greenhouse gases and roughly 30\% of anthropogenic CO$_2$ emissions~\cite{glorys12}.
Understanding the three-dimensional structure of ocean temperature and salinity, known as the thermohaline structure, is essential for monitoring ocean heat content, tracking water mass formation and circulation, predicting extreme events, and constraining climate projections.

\added{Satellite remote sensing of the ocean is inherently limited to the surface, but offers near-global coverage of variables such as sea surface temperature (SST), sea surface salinity (SSS), and sea surface height (SSH) at high spatial and temporal resolution.}
% While satellite remote sensing observes the ocean surface, providing near-global coverage of variables such as sea surface temperature (SST), sea surface salinity (SSS), and sea surface height (SSH) at high spatial and temporal resolution, these observations are inherently limited to the surface.
% This leaves anything but the very top layer
% The subsurface ocean, which constitutes practically all of the ocean's water volumes, remains sparsely observed.
The Argo program~\cite{argo2020}, with its fleet of approximately 4000 autonomous profiling floats, provides the most systematic sampling of subsurface temperature and salinity in the upper 2000\,m, but its spatial coverage still is extremely sparse relative to satellite observations and to the ocean's surface: the Core float array targets a nominal 3$^\circ \times$ 3$^\circ$ spacing. 
% \textcolor{red}{TODO: Add here ARGO->EN4 description?}
% \added{The Argo program~\cite{argo2020}, with approximately 4000 active profiling floats,
% provides the most systematic sampling of subsurface temperature and salinity in the
% upper 2000\,m. Its Core array is designed for a nominal 3-degree spacing, equivalent
% to roughly one float per \(10^5\,\mathrm{km}^2\) (a \(320\,\mathrm{km}\times320\,\mathrm{km}\)
% area-equivalent square), and thus remains spatially sparse relative to satellite
% observations and the ocean surface.}

\begin{figure}[t]
    \centering
    \includegraphics[width=0.9\linewidth]{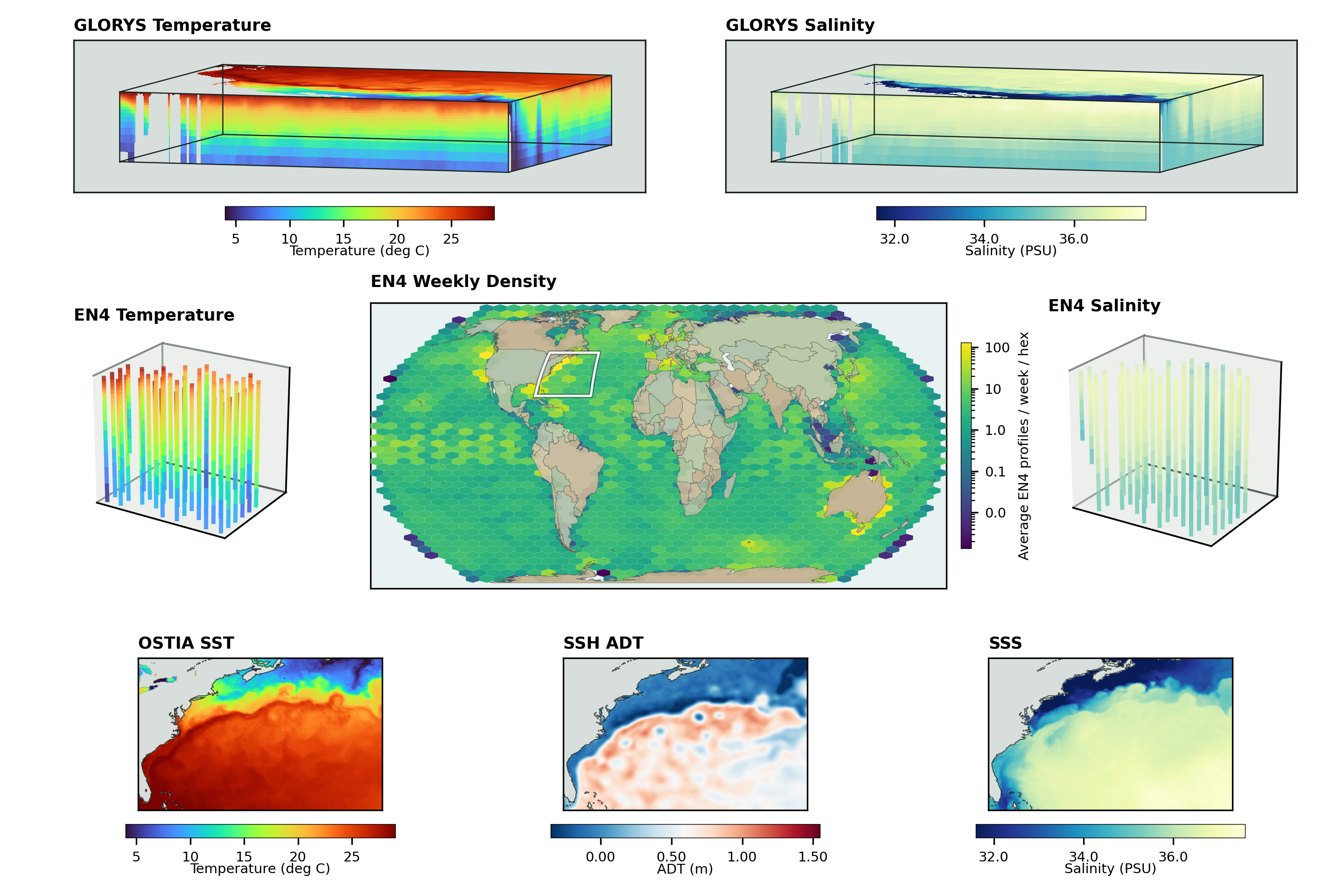}
    \caption{\textbf{Global distribution of the EN4 density per week and examples of surface observations and reanalysis data.} \datasetname aggregates all profiles gathered between 2000 and 2024 with sea surface temperature, salinity and height observational products accompanied by the GLORYS12V1 reanalysis product. The SST, ADT, SSS, GLORYS12, and profile T and S are shown for the white square around New York on the map in the middle.}
    \label{fig:argo_map}
\end{figure}

This asymmetry between dense satellite surface observations and sparse \emph{in situ} subsurface measurements has motivated research into \added{physics-based~\cite{wang2013reconstructing} or} data-driven surface-to-subsurface reconstruction where the goal is to infer the vertical thermohaline structure from satellite-observed surface fields~\cite{su2021bilstm, meng2021reconstruction, dors2022, tian2022salinity, cnn_atlantic2023, sammartino2025physically, ffpgnet2025, tscast2025, miranda2025nespreso}.
Another research direction that has recently gained traction is spatiotemporal forecasting of ocean states with deep learning ~\cite{oceanforecastbench2025, oceanbench2025, cui2025forecasting, gao2026neuralom, agarwal2026skillful}. 
Recent developments in both directions however, largely rely on reanalysis data or data generated by physical simulations.
The few exceptions to this pattern are limited in scope to specific basins, like the Gulf of Mexico or the Mediterranean.
The AI and oceanography communities would clearly benefit from a large, standardized, global dataset, but the current available datasets have the following limitations.
% Both directions benefit from globally consistent, high-resolution datasets that pair surface and subsurface observations, yet progress has been hampered by several key limitations in the available data landscape:

\begin{enumerate}[leftmargin=*, itemsep=2pt]
    \item \textbf{No standardized paired dataset.}
    Most existing works construct their own training data using ad hoc procedures, without making it publicly available.
    % by collocating satellite products with Argo profiles, using ad hoc procedures that vary across studies.
    This makes direct comparison of methods difficult and hinders reproducibility.

    \item \textbf{Reconstructed products, not observation pairs.}
    The published datasets that do exist~\cite{dors2022, tian2022salinity, liu2024seeing} are typically reconstructed gridded fields, \ie, the \emph{output} of a model, rather than the raw paired observations that could serve as training data for new methods.

    \item \textbf{Limited variable coverage.}
    Datasets cover only temperature~\cite{dors2022} or only salinity~\cite{tian2022salinity}, and similarly, surface inputs are often restricted to one or two variables rather than the full complement of SST, SSS, and SSH.

    \item \textbf{Regional scope.}
    Many deep learning studies are restricted to specific basins: the South China Sea~\cite{ffpgnet2025}, the Northwestern Pacific~\cite{tscast2025}, the Gulf of Mexico~\cite{miranda2025nespreso}, or the Atlantic~\cite{cnn_atlantic2023}.
    % Global studies exist but at coarse resolution.

    \item \textbf{Insufficient resolution.}
    Existing global datasets operate at 0.25$^\circ$ to 1.4$^\circ$ spatial resolution and monthly temporal resolution~\cite{dors2022, tian2022salinity, oceanforecastbench2025}, which is too coarse to resolve the mesoscale dynamics at $\mathcal{O}$(10--100\,km) that dominate ocean variability in many regions.

    \item \textbf{Over-reliance on reanalyses.}
    In the absence of standardized observational datasets, many modeling efforts resort to ocean reanalyses such as GLORYS12~\cite{glorys12} 
    % or objective analyses such as ARMOR3D~\cite{armor3d} 
    % both for training data and as evaluation reference.
    for both training and/or evaluation.
    For instance, benchmarks derive their inputs and ground truth entirely from GLORYS12~\cite{oceanforecastbench2025,oceanbench2025}, 
    and AI models~\cite{adaf_ocean2025,cui2025forecasting,gao2026neuralom, agarwal2026skillful} are trained against the same reanalysis data.
    % , and Su \etal~\cite{su2021bilstm} use Argo gridded products---themselves smoothed analyses---as both input and target.
    While reanalyses provide convenient dense fields, they are model outputs that inherit systematic biases from the underlying numerical model and assimilation scheme.
    % , including excessive diffusion, imperfect boundary forcing, and spurious correlations. % introduced by the smoothing of sparse observations onto a model grid.
    Models trained and evaluated exclusively against such targets risk learning to reproduce errors
    % these artifacts 
    rather than the true ocean state.
    % , and performance metrics computed against reanalysis can paint an overly optimistic picture that does not transfer to real observations.

    \item \textbf{Heterogeneous grids.}
    The application of many computer vision-inspired methods would be simplified significantly if data were on a common grid in space and time, but current datasets do not provide this~\cite{oceantaco2026}. 
    % We apply these normalizations in order to immediately generate tensors, ready to be fed into machine learning pipelines.
\end{enumerate}
To address these limitations, we introduce \datasetname, visually introduced in \Cref{fig:argo_map}: a global, high-resolution, AI-ready dataset of paired subsurface and surface ocean observations.
\datasetname pairs L4 products (optimal interpolations of satellite observed) SST, SSS, and SSH\footnote{We provide absolute dynamic topography (ADT), which is derived from sea surface height (SSH) as measured by altimeters but referenced to the Earth's geoid.} with co-located subsurface temperature and salinity profiles at \SI{0.1}{\degree}$\times$\SI{0.1}{\degree} spatial resolution and weekly temporal resolution, spanning 2000--2024.
% Each observation is additionally matched with the corresponding GLORYS12 ocean reanalysis~\cite{glorys12} profile.
\added{We include dense subsurface reanalysis data from GLORYS12~\cite{glorys12} at the same spatial and temporal resolution, which is meant to complement the observational data, not replace it.}
\added{We see the global paired subsurface--surface \textit{observations} spanning almost 25 years as the main strength of \datasetname.}
We use profiles from the EN4 dataset, which compiles profiles from different sources including Argo~\cite{en4}. 
% In our dataset, 
Profiles are interpolated to the depth levels of GLORYS12, putting them on a common depth grid and enabling direct comparison between observational and model-based targets.
% \added{This depth grid provides is limited but will }
The weekly cadence and global coverage make \datasetname suitable not only for vertical reconstruction but also for modeling ocean dynamics, like in (sub)seasonal forecasting or climate studies.

Beyond its value for ocean science, \datasetname presents distinctive challenges and opportunities for the AI and computer vision communities.
The global gridded data at \SI{0.1}{\degree} resolution provides a large-scale, real-world testbed for spatial generative and predictive models.
The 25-year weekly record (1283 time steps) enables the development of methods that capture medium to long-term temporal dynamics.
The data is inherently four-dimensional (latitude, longitude, depth, and time) and
% , posing scalability challenges for architectures that must reason over large spatial volumes; the data 
can be processed as local patches or globally, inviting research on multi-scale approaches.
Finally, and most distinctively, the subsurface observations are \textit{extremely sparse}: with a median of 13--16 Argo profiles per 128$\times$128 patch, the per-depth-level observation rate is just over 0.01\%, corresponding to $\sim$99.9\% missing data.
This far exceeds the sparsity regimes explored by current methods for learning from incomplete observations~\cite{daras2023ambient,zhou2025incomplete}.
% by 1-2 orders of magnitude
% : Ambient Diffusion~\cite{daras2023ambient} assumes access to partially masked training images, and recent diffusion-based imputation methods~\cite{zhou2025incomplete} demonstrate results down to ${\sim}$1\% observation rates---an order of magnitude denser than the Argo sampling in our dataset.
Handling such extreme and spatially irregular sparsity will require novel architectures, such as graph- or mesh-based approaches that operate directly on irregularly sampled data~\cite{alexe2024graphdop,garnier2025meshmask}, or new training objectives that can learn from near-empty grids.
% \datasetname thus serves as a challenging benchmark at the frontier of sparse-data methods in computer vision.
Our contributions are summarized as follows:
\begin{enumerate}
    \item We present an open, global, AI-ready dataset of paired satellite surface observations and \emph{in situ} subsurface profiles at 0.1$^\circ$ resolution and weekly frequency, with 9.5 million profiles across 1283 weekly time steps (\cref{sec:dataset}).
    % \item We give an overview of existing datasets of ocean variables (\cref{sec:related_work}).
    % for ocean observation and reconstruction (\cref{sec:related_work}).
    \item We define a standardized evaluation protocol and provide baseline results for a representative task: subsurface ocean state reconstruction (\cref{sec:evaluation}).
    % two representative tasks---subsurface reconstruction and surface forecasting---demonstrating the dataset's versatility (\cref{sec:evaluation}).
    % \item We establish baseline results for both tasks using several deep learning and classical approaches (\cref{sec:results}).
    \item We release all data, code, and baselines as open-source resources for the oceanography and the AI communities. The code consists of reusable data export and preprocessing scripts, as well as a configurable patching system and \texttt{pytorch} DataLoaders.
\end{enumerate}
% \footnote{\url{https://to.fill.when.accepted.org}}

%% file: sec/2_related.tex
\section{Related Work}
\label{sec:related_work}

% We review existing datasets for ocean surface-to-subsurface reconstruction, organized by whether they publish open datasets or only use internally constructed training data.
% \Cref{tab:datasets} provides a systematic comparison of all datasets, while \cref{tab:methods} summarizes the methodological landscape.

% \subsection{Published Datasets}
% \label{sec:related_datasets}

\Cref{tab:datasets} shows an overview of existing datasets along with \datasetname with existing datasets. 
% The neWe discuss related work in more detail in the next paragraphs.

\begin{table*}[ht!]
\centering
\caption{\textbf{Comparison of existing public \textit{global} ocean subsurface--surface datasets.}
% ``Type'' indicates whether the dataset provides raw observation pairs (Obs.), reconstructed gridded fields (Recon.), or reanalysis/forecast products (Rean.).
Variables: T = temperature, S = salinity, U/V = currents, H = sea surface height or ADT (absolute dynamic topography), SST = sea surface temperature, SSS = sea surface salinity, W = wind, B = biological variables, nutrients, oxygen, SR/TR = spatial/temporal resolution G = Global. $^\dagger$: Preserves native sensor resolutions (2\,km--0.25$^\circ$) without resampling to a common grid.
\label{tab:datasets}}
% todo remove type and coverage
\setlength{\tabcolsep}{3pt}
\begin{tabular*}{\linewidth}{@{\extracolsep{\fill}}lccccccc@{}}
\toprule
\textbf{Dataset}& \textbf{Surface} & \textbf{Subsurf.} & \textbf{Depth} (m) & \textbf{Levels} & \textbf{SR} & \textbf{TR} & \textbf{Period} \\
\midrule
\multicolumn{8}{@{}l}{\emph{Interpolation / Analysis}} \\
ARMOR3D~\cite{armor3d}  & -- & T, S & 1500 & 50 & 1/8$^\circ$ & D & 1993--...  \\
% Interpolated Argo TODO  & -- & -- & 0.5$^\circ$ & -- & 2002--2020 & Global \\
ISAS~\cite{isas}  & -- & T, S & 5500 & 187 & 0.5$^\circ$ & M & 2002--2020  \\
WOA ~\cite{locarnini2024world} & -- & T, S, B & 5500 & 102 & 0.25--1$^\circ$ & M & 1971--2022   \\
IAPv4~\cite{cheng2024iapv4} & -- & T & 6000 & 119 & 1$^\circ$ & M & 1940--2023 \\
RG-Clim~\cite{roemmich20092004} & -- & T, S & 2000 & 58 & 1$^\circ$ & M & 2004--2018 \\
\midrule
\multicolumn{8}{@{}l}{\emph{Reanalysis}} \\
GLORYS12~\cite{glorys12} & SST, H, Ice & T, S, U, V & 5728 & 50 & 1/12$^\circ$ & D & 1993--...   \\
% CORA \\
OceanFcstB.~\cite{oceanforecastbench2025} & SST, H, W & T, S, U, V & 650 & 23 & 1.41$^\circ$ & D & 1993--2020   \\
OceanB.~\cite{oceanbench2025} & SST, H, W & T, S, U, V & 5728 & 50 & 1/12$^\circ$ & D & 2024   \\
\midrule
\multicolumn{8}{@{}l}{\emph{Reconstruction}} \\
DORS~\cite{dors2022} & -- & T & 2000 & 23 & 1$^\circ$ & M & 1993--2020  \\
DORS0.25$^\circ$~\cite{dors025_2024} & -- & T & 2000 & 23 & 0.25$^\circ$ & M & 1993--2023 \\
IAP Salinity~\cite{tian2022salinity} & -- & S & 2000 & 41 & 0.25$^\circ$ & M & 1993--2018  \\
Liu 2024~\cite{liu2024seeing} & -- & T, S & 2000 & 187 & 0.25$^\circ$ & M & 2012--2020  \\
\midrule 
\multicolumn{8}{@{}l}{\emph{Observational}} \\
Argo~\cite{argo2020} & -- & T, S, U, V, B  & 2000 & -- & Pt. & -- & 1999--...  \\
WOD~\cite{mishonov2024world} & -- & T, S, B & 7800 & -- & Pt. & -- & 1772--2022  \\
EN4~\cite{en4} & -- & T, S  & -- & 400 & Pt. & -- & 1900--...  \\
Aquarius--A.~\cite{nasa_aquarius_argo} & SSS & S & 10 & 2 & Pt. & -- & 2011--2015  \\
OceanTACO~\cite{oceantaco2026} & SST, SSS, H, W & T, S & 2000 & -- & Nat.$^\dagger$ & D & 2015--2025  \\
\midrule
\textbf{\datasetname} & \textbf{SST, SSS, H} & \textbf{T, S} & 5728 & 50 & \textbf{0.1$^\circ$} & \textbf{W} & \textbf{2000--2024} \\
\bottomrule
\end{tabular*}
\end{table*}

\paragraph{Subsurface profiles.} 
% TODO Subsurface temperature and salinity profiles are sourced from t
The Argo program~\cite{argo2020} maintains a global fleet of ${\sim}$4000 autonomous profiling floats that measure temperature and salinity profiles.
Each float descends to a parking depth (typically 1000\,m), drifts with the currents, and periodically descends to a target depth (2000\,m or for \added{the small subset of Deep Argo} floats; 6000\,m) before ascending to the surface while measuring temperature and salinity at multiple depth levels.
In addition \added{to CTD measurements like those provided by Argo floats}, a variety of subsurface measurement techniques exist, such as MBT, XBT, or CDT~\cite{abraham2013review}.
Datasets such as EN4 (UK Met Office)~\cite{en4} and World Ocean Database (WOD, by NOAA)~\cite{mishonov2024world} bundle such subsurface measurements from various sources and apply bias corrections and quality control to different extents.
Raw profiles are measured at irregular depth levels that vary between floats and individual casts.

\paragraph{Reanalysis and analysis products.}
Global ocean reanalyses and objective analyses provide spatially dense, dynamically consistent 3D fields and are widely used as training targets or evaluation references in AI for oceanography.
GLORYS12~\cite{glorys12} is a global reanalysis at (1/12$^\circ$, daily) 
% produced by Mercator Oc\'ean International, 
that assimilates satellite altimetry, SST, sea ice, and in situ profiles through a variational data assimilation system coupled to the NEMO ocean model.
% ARMOR3D~\cite{armor3d} combines satellite altimetry and SST with in situ profiles via statistical projection to produce weekly global 3D temperature and salinity fields at (1/8$^\circ$, daily).
% The In Situ Analysis System (ISAS)~\cite{isas} provides gridded temperature, salinity, and dissolved oxygen fields from Argo optimal interpolation at (0.5$^\circ$, monthly) with 187 depth levels,
% % While valuable for oceanographic analysis, ISAS 
% but does not include satellite surface observations.
% The World Ocean Atlas~\cite{locarnini2024world} TODO
% The Multi-Mission OI SSS product~\cite{multimission_sss} concatenates satellite SSS from Aquarius, SMAP, and SMOS at 0.25$^\circ$ 4-daily resolution from 2011 to present, but again covers only the surface.  % dropped because surface only
Objective analyses such as ARMOR3D~\cite{armor3d}, the In Situ Analysis System (ISAS)~\cite{isas}, the World Ocean Atlas (WOA) ~\cite{locarnini2024world}, the Institute of Atmospheric Physics' product (IAPv4)~\cite{cheng2024iapv4} \added{or the Roemmich-Gilson Argo Climatology (RG-Clim)~\cite{roemmich20092004}} combine profiles with optimal interpolation or statistical methods into a dense grid.
While these products offer the appealing property of gap-free global coverage, they are model outputs rather than direct observations.
Reanalyses inherit systematic biases from the underlying numerical model (e.g., diffusive mixing, imperfect boundary conditions) and from the assimilation scheme, which smooths observations onto the model grid and can introduce spurious correlations.
% GLORYS12 is a reanalysis rather than an independent reference: its system assessment 
\added{For instance, GLORYS12~\cite{glorys12} reports a residual seasonal temperature bias above 100~m and 
% salinity errors that are larger before the Argo-era expansion of in-situ coverage 15, Sec. 3.1. I
independent observations along the 59.5$^\circ$ N Atlantic section found 
% good agreement for upper-700 m heat content, but 
significant differences in heat content at 700--2000~m and in overflow waters~\cite{verezemskaya2021assessing}}.
Analysis products rely on statistical relationships that may not hold everywhere, such as in dynamically complex regions, and that over-smooth variables.
% Consequently, models trained exclusively against reanalysis targets risk learning to reproduce these artefacts rather than the true ocean state, and evaluation against reanalysis alone can mask errors that would be apparent against independent observations.
Our dataset addresses this by providing both raw and matched EN4 profiles together with GLORYS12 and dense surface observations, enabling users to train and evaluate against either target and to quantify the discrepancy directly.

\paragraph{Reconstructed gridded products.}
Several works have produced global or regional gridded datasets of subsurface ocean variables derived from satellite observations and/or subsurface profiles, using machine learning or statistical methods.
~\cite{wang2013reconstructing}
The Deep Ocean Remote Sensing (DORS) dataset~\cite{dors2022} provides global subsurface temperature at 23 depth levels 
% (30--2{,}000\,m) 
from 1993--2020, reconstructed using ConvLSTM networks from satellite SST, absolute dynamic topography (ADT), and sea surface wind fields combined with EN4 profiles, at 1$^\circ \times$1$^\circ$ monthly resolution.
A higher-resolution version, DORS0.25$^\circ$~\cite{dors025_2024}, extends the record to 2023 at 0.25$^\circ$ using Deep Forest models.
Tian \etal~\cite{tian2022salinity} published a global subsurface salinity dataset at 0.25$^\circ$ and monthly resolution for 1993--2018,
% (1--2{,}000\,m, 1993--2018), 
reconstructed with MLPs from satellite ADT, SST, and sea surface wind combined with coarse gridded salinity.
Liu~\cite{liu2024seeing} proposed a physics-informed reconstruction of upper-ocean temperature and salinity 
% (surface to 400\,m) 
at 0.25$^\circ$.
% , though this work was not accepted after peer review.
% A critical distinction is that a
All of the mentioned products are \emph{reconstructed products}, i.e., model outputs that have been gap-filled and smoothed, rather than raw paired observations suitable for training new models.
% They cannot serve as ground truth or as standardized ML training data.

\paragraph{Ocean forecasting benchmarks.}
Recent efforts have produced AI-oriented benchmarks for ocean prediction, but are heavily based on reanalysis data.
Most notably, OceanForecastBench (OceanFcstB.) ~\cite{oceanforecastbench2025} provides a training dataset derived from 
% 13\,TB of 
GLORYS12 data, covering 4 subsurface variables across 23 depth levels and 4 surface variables, regridded to 1.41$^\circ$ resolution for 1993--2020.
OceanBench (OceanB.) ~\cite{oceanbench2025} defines evaluation tracks with observational data for short-range forecasting using GLORYS12, ERA5 and physical model forecasts as training data.
% \added{(exclusively for the year 2024)}
Both benchmarks include observational data but only for evaluation and in smaller quantities: e.g., OceanBench only includes \added{observational data such as profiles for the year 2024, and OceanForecastBench only for 2022-2023, while \datasetname includes observations spanning 2000--2024. OceanForecastBench is additionally limited by its coarse spatial resolution of 1.41$^\circ$ (compared to 0.1$^\circ$ in \datasetname}. 
% and only provide reanalysis as training data. 
% and uses reanalysis data from GLORYS12 and ERA5.

% While these benchmarks are significant for the ocean ML community, 
% they target \emph{temporal forecasting} (predicting future ocean states) rather than \emph{vertical reconstruction} (inferring subsurface structure from surface observations), and their coarse resolutions preclude mesoscale analysis.

% \paragraph{In situ gridded products.}
% The In Situ Analysis System (ISAS)~\cite{isas} provides gridded temperature, salinity, and dissolved oxygen fields from Argo optimal interpolation at 0.5$^\circ \times$ 0.5$^\circ$ with 187 depth levels.
% While valuable for oceanographic analysis, ISAS does not include satellite surface observations and is not structured for ML workflows.

\paragraph{Multi-sensor collections.}
In a parallel work, OceanTACO~\cite{oceantaco2026} (in review, only preprint available) provides a harmonized, global multi-sensor sea state dataset spanning 2015--2025 (vs. 2000--2024 for \datasetname), integrating satellite altimetry, SST, SSS, surface winds, GLORYS12 reanalysis, and Argo~\cite{argo2020} in situ profiles under a unified cloud-optimized specification at daily temporal resolution.
OceanTACO, \added{like OceanBench and OceanForecastBench, only propose Argo profiles as an evaluation resource, and consequently, they include only 400K profiles gathered between 2023 and 2025 (vs. 9.5M in \datasetname, for 2000-2024)}.
% only uses Argo profiles and not a compilation dataset like EN4, resulting in a smaller number of profiles (only 140K instead of our 9.5M, which they propose as evaluation resource rather than as for training, in contrast to our goal in this work), and they provide a shorter historical record.
OceanTACO 
% encompasses more data sources than \datasetname and preserves native observation characteristics, but 
does not resample spatial resolutions across modalities to a common grid, does not interpolate depth levels onto a standard vertical coordinate, and does not provide a patching system that tiles the globe into fixed-size tensors directly ingestible by deep learning models. 
This makes \datasetname more into a more AI-ready resource. 
% We also provide a longer historical record.
% This makes OceanTACO a comprehensive and configurable starting point for building task-specific datasets, whereas \datasetname is designed to be immediately AI-ready: downloadable and loadable by a \texttt{pytorch} data pipeline without additional preprocessing.
% OceanTACO is currently a preprint under review at Earth System Science Data, which targets the Earth science community, while \datasetname is aimed at the AI and computer vision communities.
% The parallel OceanTACO~\cite{oceantaco2026} dataset offers a broader collection of surface sources at daily resolution, but preserves native sensor resolutions without resampling to a common grid, interpolating depth levels, or providing a patching system.
% \datasetname is designed for immediate use in deep learning pipelines: all data is re-gridded to a uniform \SI{0.1}{\degree} grid, depth-interpolated onto 50 standard levels, and pre-tiled into 128$\times$128 (configurable) patches loadable directly by a \texttt{pytorch} data loader in a matter of minutes.
NASA's Aquarius--Argo (Aquarius-A.) validation dataset~\cite{nasa_aquarius_argo} provides collocated satellite SSS and Argo surface measurements for the Aquarius mission period (2011--2015), but only extends up to 10m of depth.

% \subsection{Comparison with Existing Datasets}
% \label{sec:comparison}

\paragraph{}
The key features of our dataset are summarized as follows.

\begin{enumerate}[leftmargin=*, itemsep=2pt]
    \item \textbf{Paired observations.}
    Unlike other sources~\cite{glorys12,tian2022salinity,dors2022,dors025_2024} which consist of reconstructions or reanalyses,
    % DORS~\cite{dors2022}, DORS0.25$^\circ$~\cite{dors025_2024}, and the IAP salinity dataset~\cite{tian2022salinity}, which publish reconstructed gridded fields (model outputs), 
    \datasetname provides raw paired observations for the development of observation-based AI methods.

    \item \textbf{High spatial resolution.}
    At \SI{0.1}{\degree}, our dataset is finer than all global datasets except OceanTACO~\cite{oceantaco2026} where variables are not on a common grid.
    % (DORS0.25$^\circ$) and 14$\times$ finer than OceanForecastBench (1.4$^\circ$).

    \item \textbf{Weekly temporal resolution and historical record of 25 years.}
    % Most existing datasets are monthly; \datasetname provides weekly data, better capturing intra-monthly variability.
    \added{Existing datasets like OceanTACO, OceanBench and OceanForecastBench only provide subsurface profiles as an evaluation resource and only for 1-2 years.}

    \item \textbf{Coverage of variables.}
    % TODO
    % DORS provides only temperature; the IAP dataset provides only salinity.
    % All three surface variables (SST, SSS, SSH) as inputs.
    SST, SSS, SSH (ADT) are provided as surface variables, in addition to temperature and salinity as subsurface variables.

    \item \textbf{Paired reanalysis.}
    The inclusion of matched dense EN4 profiles on the GLORYS12 reanalysis hypercubes alongside the surface observations is unique and enables curriculum or multi-stage learning, and immediate comparison. 
    % Building training-ready tensors from this data shape is trivial.

    \item \textbf{Global geographical coverage.}
    % Unlike regional studies (TS-Cast: NW Pacific; NeSPReSO~\cite{miranda2025nespreso}: Gulf of Mexico; FFPG-net: South China Sea; CNN~\cite{cnn_atlantic2023}: Atlantic), \datasetname covers all oceans.

    \item \textbf{AI-ready format.}
    Data is on a common grid and a configurable patching system is provided, supporting easy loading as tensors.
    
\end{enumerate}

%% file: sec/3_dataset.tex
\section{The \datasetname Dataset}
\label{sec:dataset}

 \begin{table}[t]                                                                                                                                                                                                         
    \centering
    \caption{Dataset summary statistics.}                                                                                                                                                                                
    \label{tab:dataset-summary}
    \begin{tabularx}{\linewidth}{@{}>{\raggedright\arraybackslash}p{0.29\linewidth}X@{}}
      \toprule
      \textbf{Property} & \textbf{Value} \\
      \midrule
      Spatial coverage & Global, 3{,}600$\times$1{,}800 px at \SI{0.1}{\degree} resolution (4{,}338{,}138 ocean cells) \\
      Temporal coverage & January 2000 -- July 2024, weekly resolution (1283 target weeks) \\
      Total Argo profiles & 9{,}485{,}977 (9.4\,M valid temperature; 7.0\,M valid salinity) \\
      Surface variables per sample & SST (OSTIA), SSS \& surface density (MULTIOBS), ADT (DUACS) \\
      Depth levels & 50 GLORYS standard levels (0.49\,m to 5{,}728\,m) \\
      Patch dataset & 358\,K patch--date samples ($128^2$ px, no overlap) -- 343\,K train / 15\,K eval \\
      Median profiles per patch--date & 13--16 \\
      Source files & 169 EN4/Argo, 843 GLORYS, 5{,}326 OSTIA, 5{,}326 sea-level, 5{,}326 SSS \\
      Storage format & Zarr (profiles) + GeoTIFF/ZSTD (dense rasters) + Parquet (indices) \\
      \bottomrule
    \end{tabularx}
  \end{table}

\Cref{tab:dataset-summary} summarizes the key properties of \datasetname.
This section describes the data sources, construction pipeline, storage format and patching system of \datasetname.
All data export and processing scripts, along with \texttt{pytorch} DataLoader code, 
% can be found in an anonymized repository and 
which are made publicly available \added{under a permissive \texttt{CC BY 4.0} license}: \url{https://huggingface.co/datasets/ESA-philab/OceanDepths}.

\added{We are not currently planning to update our dataset continuously as new data becomes available, we provide our dataset as a static resource to be used for training and intercomparable evaluation by the ML community.}

\subsection{Data Sources}
\label{sec:data_sources}

\datasetname integrates three complementary data sources: subsurface observations, remotely sensed surface observations, and a dense reanalysis.

\subsubsection{Subsurface Profiles.}
\label{sec:argo}

% TODO Subsurface temperature and salinity profiles are sourced from the Argo program~\cite{argo2020}, which maintains a global array of ${\sim}$4{,}000 autonomous profiling floats.
% Each float descends to a parking depth (typically 1{,}000 or 2{,}000\,m), drifts with the currents, and periodically ascends to the surface while measuring temperature and salinity at multiple depth levels.

We use quality-controlled profiles from the EN4.2.2\footnote{\href{https://www.metoffice.gov.uk/hadobs/en4/download-en4-2-2.html}{\url{https://www.metoffice.gov.uk/hadobs/en4/download-en4-2-2.html}}.}\hfill\break archive~\cite{en4},
% maintained by the UK Met Office Hadley Centre, 
which integrates Argo~\cite{argo2020} float profiles with ship-based CTD and other hydrographic measurements such as XBT~\cite{abraham2013review}.
Each profile provides 
% \added{potential} 
temperature (\texttt{TEMP}) and bias-corrected salinity (\texttt{PSAL\_CORRECTED}) at instrument-specific corrected depths (\texttt{DEPTH\_CORRECTED}), with up to 400 depth samples per profile.
Raw profiles are measured at irregular depth levels that vary between floats and individual casts.

\subsubsection{Satellite Surface Observations.}
\label{sec:satellite}

We use gridded satellite products for three surface variables that jointly constrain the upper-ocean state.
% thermohaline structure. 
In order to have continuous and uniform information, we assume that L4 data products are close enough to raw observations (no physical model or assimilation used)\footnote{\added{Interpolated L4 products nevertheless introduce some of the errors inherent to analysis products described in Sec. \ref{sec:related_work}. Hence, we plan to include L3 products of surface observations, without interpolation, to \datasetname in the future.}} while providing ease of use, and we select:

\begin{itemize}[leftmargin=*, itemsep=2pt]
    \item \textbf{Sea Surface Temperature (SST):}
    The OSTIA (Operational Sea Surface Temperature and Ice Analysis) L4 product with daily gap-free SST at \SI{0.05}{\degree} native resolution, from the UK Met Office.\footnote{Product~ID: \texttt{SST\_GLO\_SST\_L4\_REP\_OBSERVATIONS\_010\_011}.}
    %, distributed via the Copernicus Marine Service.
    % OSTIA provides 
    % offering daily gap-free SST at \SI{0.05}{\degree} native resolution.
    % which we regrid to our target \SI{0.1}{\degree} grid.

    \item \textbf{Sea Surface Salinity (SSS):}
    The MULTIOBS global sea surface salinity product from the Copernicus Marine Service\footnote{Product~ID: \texttt{MULTIOBS\_GLO\_PHY\_S\_SURFACE\_MYNRT\_015\_013}.} with daily SSS and surface density fields at 0.125$^\circ$ that combines multiple satellite and in situ sources.

    \item \textbf{Sea Surface Height -- Absolute Dynamic Topography (ADT):}
    The DUACS L4 multi-satellite altimetry product\footnote{Product~ID: \texttt{SEALEVEL\_GLO\_PHY\_L4\_MY\_008\_047}.} at 0.125$^\circ$ native resolution, providing daily absolute dynamic topography (ADT) and geostrophic currents.
\end{itemize}

\subsubsection{GLORYS12 Ocean Reanalysis.}
\label{sec:reanalysis}

The GLORYS12V1 (Global Ocean Physics Reanalysis)~\cite{glorys12} is a global ocean reanalysis based on the NEMO ocean model at 1/12$^\circ$ (${\sim}$8\,km) horizontal resolution with 50 vertical levels spanning 0.49\,m to 5728\,m depth, assimilating \emph{in situ} profiles (including Argo), satellite altimetry, SST, and sea ice concentration via a reduced-order Kalman filter.
% GLORYS12 provides daily snapshots of 3D temperature (\texttt{thetao}) and salinity (\texttt{so}) fields, which serve as both the dense reconstruction target and the source of the standard vertical depth coordinate adopted throughout \datasetname.
GLORYS12 consists of daily snapshots of 3D temperature (\texttt{thetao}) and salinity (\texttt{so}) fields. 
Its depth levels are used as the standard vertical coordinate throughout 
\datasetname.
% , which we aggregate to weekly frequency, and which serve as the source of the standard vertical depth coordinate adopted throughout \datasetname.
% For each weekly time step, we extract the full GLORYS12 temperature and salinity fields on the \SI{0.1}{\degree} grid, 
% \added{GLORYS12 assimilates SST and along-track sea-level anomalies, but not satellite SSS~\cite{glorys12}; hence differences from our separately mapped OSTIA, DUACS, and MULTIOBS fields are inter-product differences, not independent estimates of GLORYS12 error.
\added{Note that although GLORYS12 assimilates satellite SST and altimeter sea-level anomalies, some disagreement between GLORYS12 and the surface observations in \datasetname can be expected due to differences in the processing systems that produced the products. 
}
% Therefore, differences from the separately mapped OSTIA SST and DUACS ADT products quantify disagreement between processing systems, rather than an independent error estimate for GLORYS12; MULTIOBS SSS provides a more independent surface comparison, although it is also a gridded analysis rather than ground truth.
GLORYS12 provides a spatially dense source that complements the point-wise profiles.
This enables users to:
\begin{itemize}[leftmargin=*, itemsep=2pt]
    \item Train on reanalysis and validate against EN4 observations, or vice versa.
    \item Train on reanalysis and finetune on Argo profiles in a second stage. 
    \item Study the discrepancies between observed and reanalysis subsurface structure.
    \item Use reanalysis as a dense spatial complement to sparse Argo sampling.
\end{itemize}

\begin{figure*}[t]
    \centering
    \includegraphics[width=0.99\textwidth]{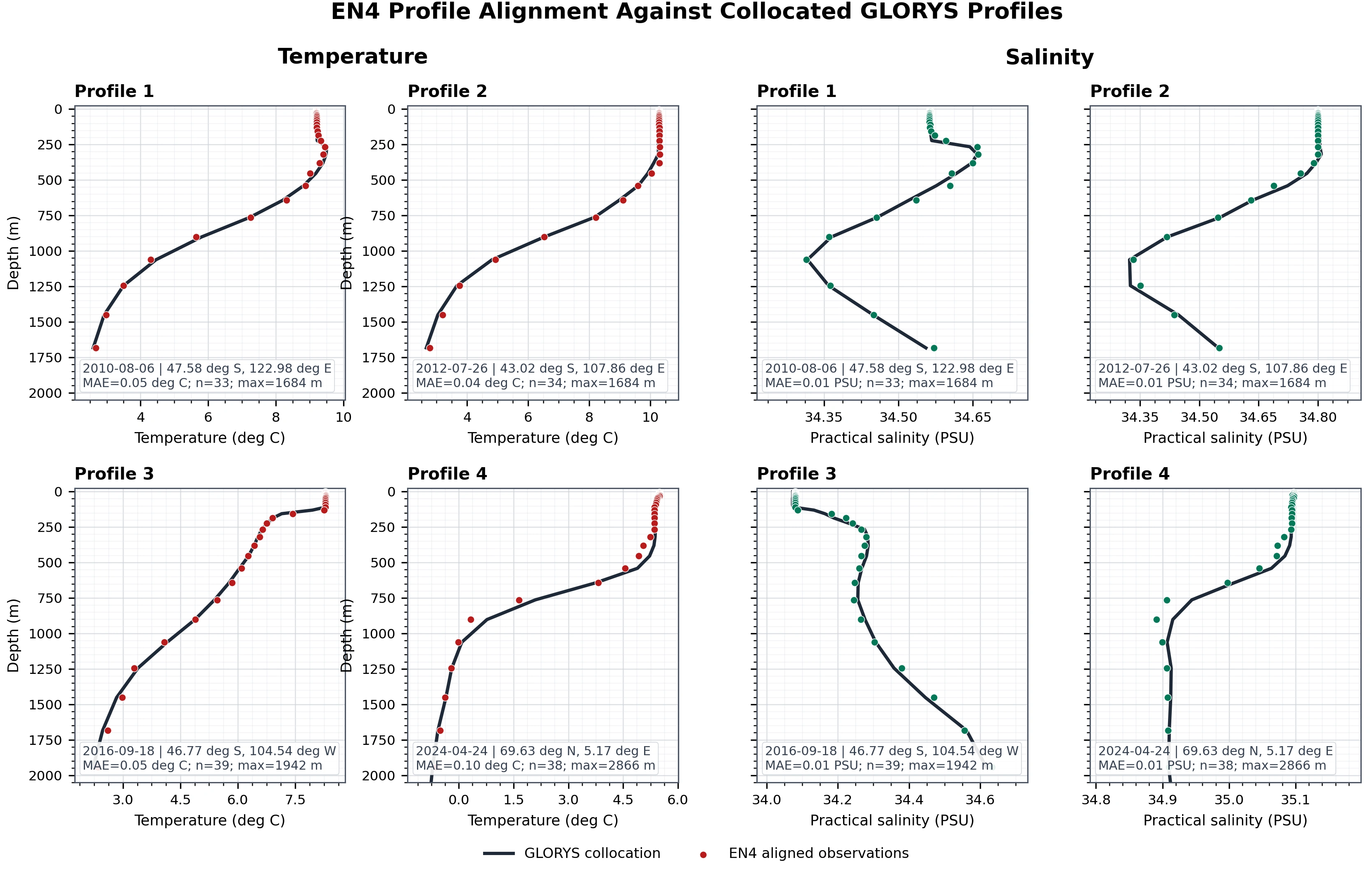}
    \caption{\textbf{EN4--GLORYS12 profile alignment example.}
    Example from the South Atlantic in September 2018. The depth-interpolated profiles closely track the GLORYS12 reanalysis across the full water column for both modalities, with occasional discrepancies (which is expected due to limitations of the reanalysis method and/or uncertainties in the profile measurements).}
    \label{fig:profile_alignment}
\end{figure*}

\subsection{Data Processing}
\label{sec:construction}

% The dataset construction involves four main steps. \cref{tab:dataset-summary} summarizes the key properties of \datasetname.

\subsubsection{Step 1: EN4 profile selection and quality control.}
We ingest all EN4 yearly archives from 2000-2024 (169 files) and extract profiles with valid temperature and/or salinity measurements. Profiles are filtered to retain only those with finite depth-temperature pairs.

\subsubsection{Step 2: Interpolating profiles to the GLORYS12 depth levels.}
% Depth alignment to the 50 GLORYS levels yields 6.6M profiles with valid temperature and 5.4M with valid salinity.
To ensure consistency, all profiles are projected onto the 50 fixed depth levels of the GLORYS12 reanalysis (\cref{sec:reanalysis}) via linear interpolation. \autoref{fig:depth_distribution} shows the distribution of EN4 index-levels with their associated depth, highlighting the fact that building training tensors from the raw dataset is not straightforward.
For a profile with finite, sorted samples $(z_i, x_i)$, where $z_i$ denotes corrected depth and $x_i$ either temperature or salinity, duplicate depths are first averaged.
For each GLORYS target depth $g_k$, we define the nearest observed profile depth
\[
z^*_k = \arg\min_{z_i} |z_i - g_k|
\]
and accept an interpolated value only if
\[
g_k \in [z_1, z_n]
\quad\text{and}\quad
|z^*_k - g_k|
\leq
\max\left(0.1\,g_k,\;10\,\mathrm{m}\right).
\]
Accepted values are computed by one-dimensional linear interpolation,
\[
\tilde{x}(g_k)
=
x_j
+
\frac{g_k-z_j}{z_{j+1}-z_j}
\left(x_{j+1}-x_j\right),
\qquad
z_j \leq g_k \leq z_{j+1}.
\]
This depth-adaptive acceptance criterion is more restrictive near the surface (where profiles are densely sampled) and more permissive at depth (where sampling is sparser).
No values are produced outside the observed depth range. All rejected target depths are marked as missing, and no extrapolation is performed outside the observed depth range.
This alignment step yields 9.5M profiles with valid temperature and 7.0M with valid salinity.
\Cref{fig:profile_alignment} shows examples of aligned EN4 and GLORYS12 profiles.

\subsubsection{Step 3: Raster product regridding and spatial profile collocation.}
All surface products and GLORYS12 are regridded onto a common \SI{0.1}{\degree}$\times$\SI{0.1}{\degree} global grid of 3600$\times$1800 pixels. We use nearest-neighbor selection for sources with matching resolution and bilinear interpolation otherwise.

% Profiles get assigned to the \SI{0.1}{\degree}$\times$\SI{0.1}{\degree} cell their GPS coordinate is nearest to.
\added{EN4 profiles are not horizontally interpolated; they are assigned only to their nearest \SI{0.1}{\degree}$\times$\SI{0.1}{\degree} cell. This produces a maximum point-to-cell-center mismatch of 7.9 km at the equator, comparable to the native GLORYS12 grid scale, and should only affect local comparisons across sharp fronts.}
% ; linear depth interpolation may additionally smooth sharp vertical gradients, but we do not extrapolate outside the observed profile range.

\subsubsection{Step 4: Temporal collocation.}
For temporal alignment, weekly SST, ADT, and SSS fields, as well as a weekly GLORYS12 cube, are computed as centered 7-day means around the weekly chosen target dates, ensuring that the surface observations are representative of the weekly period and that there is temporal coherence across products.
% at weekly resolution. 
EN4 profiles are assigned to the week (centered around the same target dates) in which the measurement was performed.

\subsubsection{Step 5: Satellite collocation and context sampling.}
 For each retained EN4 profile, we sample all surface and GLORYS12 variables from the co-located \SI{0.1}{\degree}$\times$\SI{0.1}{\degree} grid cell and save these as context to the associated profile, along with the index of the grid cell in the raster the profile got assigned to.

% OSTIA SST, SSH (ADT), and SSS fields are temporally aggregated as centered 7-day means around the corresponding GLORYS weekly target date, ensuring that satellite context is representative of the weekly period rather than a single snapshot.
% All products are regridded to the common \SI{0.1}{\degree} global grid using nearest-neighbor selection for matching-resolution sources and bilinear interpolation otherwise.

% \paragraph{Step 3: Reanalysis collocation.}
% GLORYS12 weekly temperature and salinity fields are exported as dense rasters on the same \SI{0.1}{\degree} grid.
% The 50-level depth axis provides the vertical coordinate for the entire dataset. These levels are densely spaced near the surface but sparsity increases with depth.

% Dense rasters are quantized to uint8 encoding (values 0--254 for data, 255 for nodata) with ZSTD compression.
% Temperature spans 270.15--308.15\,K (quantization error $\leq$ 0.075\,K); salinity spans 30--40\,PSU (error $\leq$ 0.020\,PSU).

% GLORYS12 provides daily snapshots of 3D temperature (\texttt{thetao}) and salinity (\texttt{so}) fields
% , which we aggregate to weekly frequency, and which serve as the source of the standard vertical depth coordinate adopted throughout \datasetname.
% For each weekly time step, we extract the full GLORYS12 temperature and salinity fields on the \SI{0.1}{\degree} grid, providing a spatially dense target or input that complements the point-wise profiles.

\subsubsection{Step 6: Data format, compression, and access.}
\label{sec:format}

% \paragraph{Step 6: Storage and compression.}
All dense rasters (surface observations and GLORYS12) are linearly quantized to 8-bit GeoTIFFs to reduce storage and I/O while retaining observational precision. 
Valid values use integer codes $0,\ldots,254$, with code $255$ reserved for \textit{nodata}, and are decoded as
\[
\hat{x} = x_{\min} + \frac{q}{254}(x_{\max}-x_{\min}).
\]
Temperatures are stored in Kelvin, introducing a $+273.15$ shift for Celsius inputs, with SST stretched over $[270.15,308.15]$ K; SSS is stretched over $[30,40]$ PSU; and ADT over $[-2,2]$ m. The stretch bounds, units, \textit{nodata} code, decode formula, quantization step, and maximum absolute quantization error are written into each GeoTIFF's metadata. Excluding any (extremely rare) clipping outside the fixed stretch range, the worst-case rounding loss is half a quantization step: \SI{0.075}{K} for temperature values, \SI{0.020}{PSU} for salinity values, and \SI{0.0079}{m} for values in meters (ADT).

\datasetname is distributed in the following three complementary formats.

\paragraph{Dense rasters (GeoTIFF).}
Spatially dense GLORYS12 fields and satellite surface products are stored as multi-band GeoTIFF files with ZSTD compression and quantized as described above. Each raster covers the full 3600$\times$1800 global grid at a single weekly time step.

\paragraph{Aligned profiles (Zarr).}
EN4 observations are stored in a Zarr archive with profile-level arrays for temperature, salinity, and validity masks projected onto the 50 GLORYS12 depth levels. These files include the collocated (in time and space) surface observations of ADT, SSS and SST, as well as GLORYS12 estimates of T, S at every depth level. Supporting parquet index files provide efficient spatiotemporal querying.

\paragraph{Raw profiles (Zarr).}
The original EN4 observations are also stored in their unaltered form, containing all measurements and quality flags for reproducibility.

\subsection{Configurable Patching and Data Splits}
\label{sec:patching}

\begin{figure}[t]
    \centering
    \includegraphics[width=\linewidth]{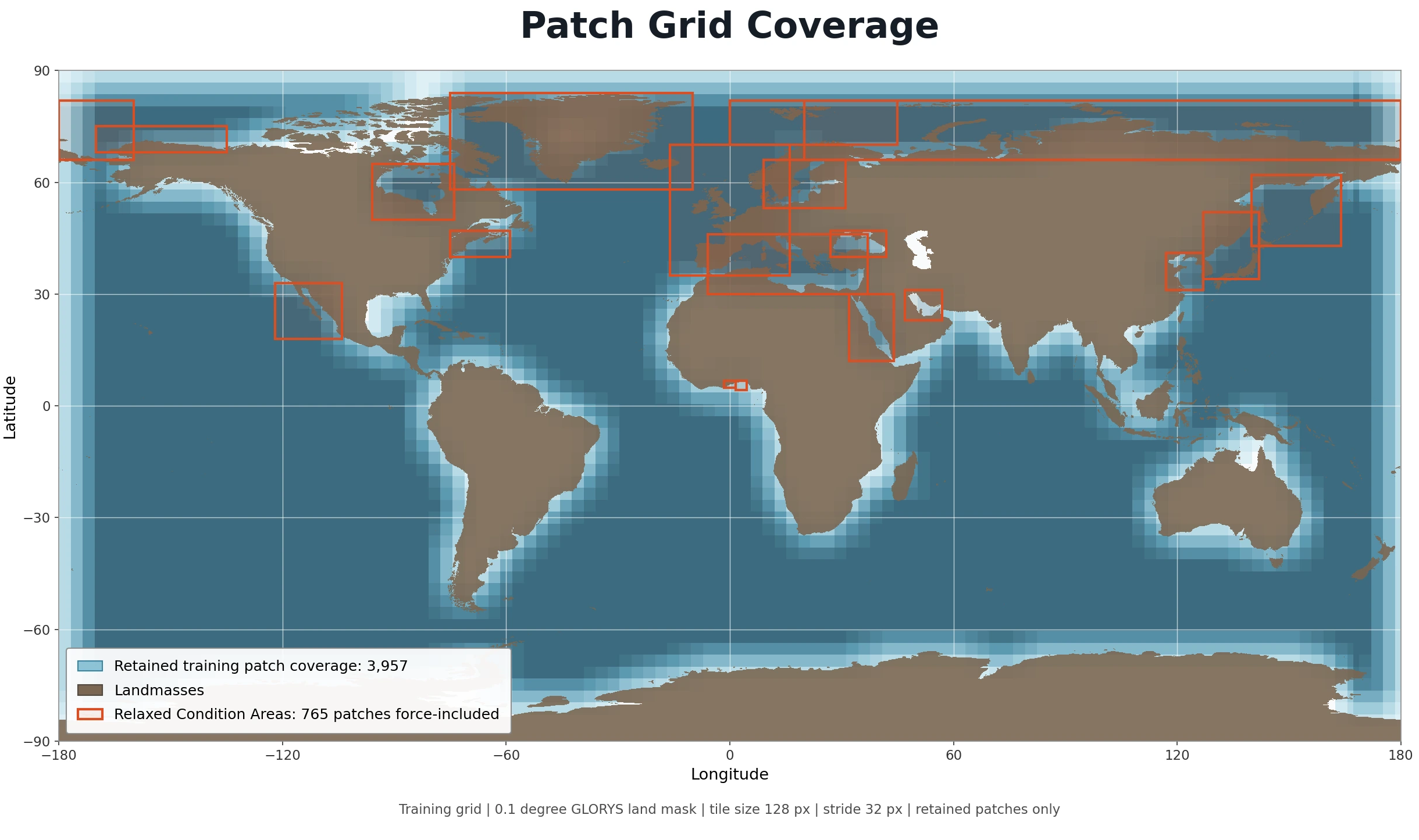}
    \caption{\textbf{Global overlapping patch grid coverage for the training dataset.} Retained patches ($\leq$30\% land, transparent blue) and force-included patches for enclosed basins (in red boxes) cover all major ocean regions. Discarded patches ($>$30\% land) are excluded. Transparency indicates overlap density from the 75\% overlap tiling.}
    \label{fig:patch_grid}
\end{figure}

\begin{figure}[t]
    \centering
    \includegraphics[width=0.8\linewidth]{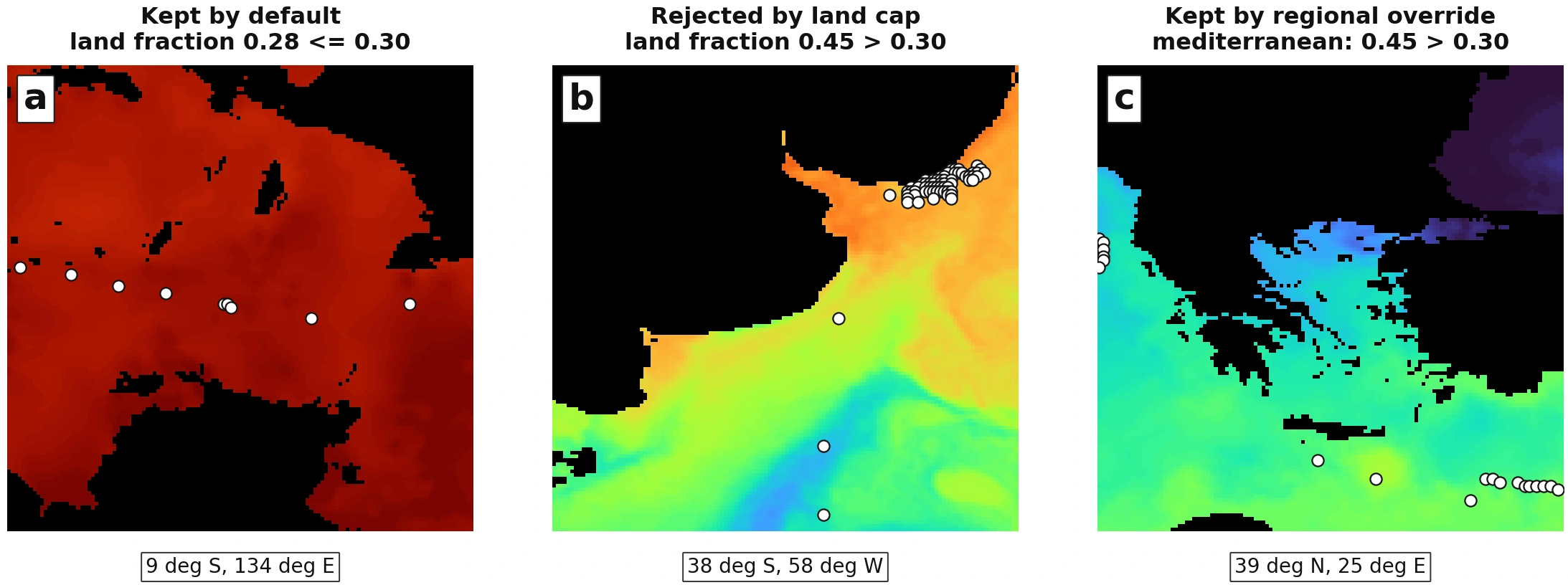}
    \caption{\textbf{Patch land-fraction filtering  with land areas (black), EN4 locations and SST.} \emph{a:} Kept by default. \emph{b:} Argentinian coast, rejected by default. \emph{c:} Aegean Sea, included due to relaxed region-based rules.}
    \label{fig:land_filter}
\end{figure}

% \paragraph{Step 7: Spatial tiling and splits.}
\datasetname comes with ready-to-use \texttt{pytorch} DataLoaders that support a variety of functions such as patched overlaps, land-pixel cutoffs, and a set of relaxed cutoff rules for certain areas (\autoref{fig:patch_grid}). We provide configuration files that allow easy changes to relevant settings (\eg, patch size). Here, we describe the recommended standard settings. See \autoref{fig:random_samples} for more examples. \newline

The patch catalogue is constructed on the GLORYS12 land-mask that has been resampled to 0.1$^\circ$. The global grid is tiled into 128$\times$128 pixel patches, corresponding to 12.8$^\circ \times$ 12.8$^\circ$, with a stride of 32 pixels (3.2$^\circ$), \ie, 75\% overlap. For each candidate patch, land fraction is computed from a binary land mask; patches with more than 30\% land are discarded. 
To retain enclosed or narrow basins of interest, that would otherwise be over-filtered, patches whose centers fall inside configured regional boxes are kept under relaxed land-fraction thresholds: Mediterranean ($\leq$60\% land), Baltic Sea ($\leq$85\%), Red Sea ($\leq$85\%), and Hudson Bay ($\leq$95\%), as illustrated in \Cref{fig:land_filter}. 
The configuration files allow to add, edit or remove such areas or constrains.
%\footnote{\url{https://anonymous.4open.science/}} 

Evaluation splits are date-based. The year 2018 is reserved for evaluation, and all other years form the training set, preventing spatial leakage between splits. The training set at $128\times128$ crop size contains 343\,K non-overlapping patches, and the evaluation set 15\,K. With the recommended strided patching, these values grow to 4.8\,M and 205\,K, respectively.
% The resulting patch dataset contains 2{,}699{,}267 total rows (2{,}214{,}599 training; 169{,}936 validation).
% \Cref{fig:patch_grid} shows the global patch coverage, and \cref{fig:land_filter} illustrates the land-fraction filtering.

%\paragraph{Training samples.}
%The data loader extracts 128$\times$128-pixel patches and returns a dictionary with keys:
%surface context \texttt{eo} $\in \mathbb{R}^{1 \times 128 \times 128}$,
%sparse observations \texttt{x} $\in \mathbb{R}^{50 \times 128 \times 128}$,
%dense target \texttt{y} $\in \mathbb{R}^{50 \times 128 \times 128}$,
%and corresponding boolean validity masks.
%Training normalization applies $\hat{T} = (T - 289.74) / 10.93$ for temperature and $\hat{S} = (S - 34.54) / 1.16$ for salinity, with missing values filled as zero after normalization.

%% file: sec/4_evaluation.tex
\section{Baseline Experiment}
\label{sec:evaluation}

We provide an example task and evaluation of baseline methods that demonstrate the use of \datasetname: the reconstruction of subsurface ocean state as given by its thermohaline structure.
% In this case we only use SST and subsurface T, but the example can easily be extended to include all variables (including SSS, SSH, subsurface S). 

%\label{sec:task_recon}

Given a 128$\times$128 spatial patch at a given weekly date, the model receives a surface context image for a given variable (either SST or SSS) $\mathbf{e} \in \mathbb{R}^{1 \times H \times W}$, sparse Argo observations for the same variable (T or S) $\mathbf{x} \in \mathbb{R}^{D \times H \times W}$ with a validity mask, and a land/ocean mask.
It predicts the dense 3D field $\hat{\mathbf{y}} \in \mathbb{R}^{D \times H \times W}$ ($D{=}50$ depth levels, $H{=}W{=}128$). 
We evaluate against the held-out EN4 profiles and against the dense GLORYS12 cube.

\subsubsection{Methods.}
We compare four baseline methods. All learning-based methods have been adapted to take SST as an additional input.
\begin{itemize}
    \item A climatology baseline computed by aggregating all EN4 training samples in each $128\times128$ patch and interpolating these by inverse-distance weighting, to approximate weekly mean observations over the 2000-2024 (excluding 2018) timeframe per location, 
    \item Nearest-profile inverse-distance weighting interpolation of the profiles in $\mathcal{P}_{\mathrm{in}}$, 
    \item A point-wise LSTM adapted from \cite{buongiorno2020deep}, 
    \item A point-wise CNN adapted from \cite{cnn_atlantic2023}, 
    \item Spatial U-Net \cite{ronneberger2015u} encoder-decoders with 2D/3D convolutions. 
\end{itemize}

\subsubsection{Metrics.}
\label{sec:metrics}

The baseline methods are evaluated using root mean squared error (RMSE), mean absolute error (MAE), and the coefficient of determination ($R^2$). These metrics are reported both at each individual depth level up to \SI{2000}{m} (\autoref{fig:res_graph}) and in a depth-integrated form \autoref{tab:recon_results}.
Since some EN4 observations are necessary for running inference for the spatial reconstruction models, we calculate the metrics on a subset of profiles. We split the available profiles $\mathcal{P}$ into an input set $\mathcal{P}_{\mathrm{in}}$ and a held-out validation set $\mathcal{P}_{\mathrm{val}}$, with
$\mathcal{P}_{\mathrm{in}} \cap \mathcal{P}_{\mathrm{val}} = \emptyset$,
$|\mathcal{P}_{\mathrm{in}}| = 0.8|\mathcal{P}|$, and
$|\mathcal{P}_{\mathrm{val}}| = 0.2|\mathcal{P}|$.
The reconstruction methods are provided only with $\mathcal{P}_{\mathrm{in}}$, and performance is evaluated exclusively at the profile locations and depth levels of $\mathcal{P}_{\mathrm{val}}$. For a prediction $\hat{{y}}_{p,z}$ and EN4 observation ${y}_{p,z}$ at profile $p$ and depth $z$, the held-out metrics are computed over all valid pairs $(p,z) \in \mathcal{P}_{\mathrm{val}}$.

%Since some EN4 profiles are required for the generation process, we hold out 20\% of the profiles and provide the model with the remaining 80\%. Performance metrics are then computed on the held-out profiles.

% \subsection{Task 2: Surface Forecasting}
% \label{sec:task_forecast}

% \textcolor{red}{Enough time to do this? We dont provide ways to return multi-temporal batches.}
% The weekly cadence and global coverage of \datasetname naturally support a temporal forecasting task.
% Given a sequence of $T$ consecutive weekly SST patches $\{\mathbf{e}_{t-T+1}, \dots, \mathbf{e}_{t}\}$, the model predicts the SST field at one or more future lead times $\hat{\mathbf{e}}_{t+\tau}$ ($\tau \in \{1,2,4\}$ weeks).
% This task tests whether methods can capture the spatiotemporal dynamics encoded in the dataset's weekly time series.

% \paragraph{Baselines.}
% We compare against four baselines: persistence using the last observed SST, weekly climatology, a ConvLSTM~\cite{shi2015convolutional}, and a U-Net that uses temporal stacking of input frames.

\subsubsection{Results.}
\label{sec:results}

% We present baseline results for both tasks on the \datasetname validation set (2018), trained on the temperature-only scenario.

% \subsubsection{Subsurface Reconstruction}
\label{sec:recon_results}

\begin{table}[t]
\centering
\caption{\textbf{Subsurface reconstruction results} (evaluation year 2018, week 25, $|\mathcal{P}_{\mathrm{val}}| = 0.2|\mathcal{P}|$).
Performance is reported for temperature and salinity against held-out EN4 profiles, averaged across depth levels no deeper than 2000 m. Best values highlighted in bold.}
\label{tab:recon_results}
\begin{tabular*}{\linewidth}{@{\extracolsep{\fill}}lcccccc@{}}
\toprule
%& \multicolumn{6}{c}{\textbf{EN4 Validation Set}} \\
\cmidrule(lr){2-7}
& \multicolumn{3}{c}{\textbf{Temperature}} 
& \multicolumn{3}{c}{\textbf{Salinity}} \\
\cmidrule(lr){2-4} \cmidrule(lr){5-7}
\textbf{Method} 
& RMSE & MAE & $R^2$
& RMSE & MAE & $R^2$ \\
\midrule
Climatology & \textbf{0.974} & \textbf{0.510} & \added{\textbf{0.964}} & 0.543 & 0.170 & 0.717 \\
IDW         & 0.979 & 0.521 & \textbf{0.964} & 0.566 & 0.174 & 0.678 \\
LSTM        & 2.420 & 1.645 & 0.701 & 0.623 & 0.295 & 0.633 \\
1D CNN      & 3.073 & 2.377 & 0.370 & 0.797 & 0.536 & 0.338 \\
3D U-Net    & 1.101 & 0.645 & 0.937 & 0.518 & 0.177 & 0.738 \\
2D U-Net    & 1.092 & 0.610 & 0.941 & \textbf{0.515} & \textbf{0.166} & \textbf{0.744} \\
% DepthDif  & -- & -- & -- & -- & -- & -- \\
\bottomrule
\end{tabular*}
\end{table}

\begin{figure}[t]
    \centering
    \includegraphics[width=0.75\linewidth]{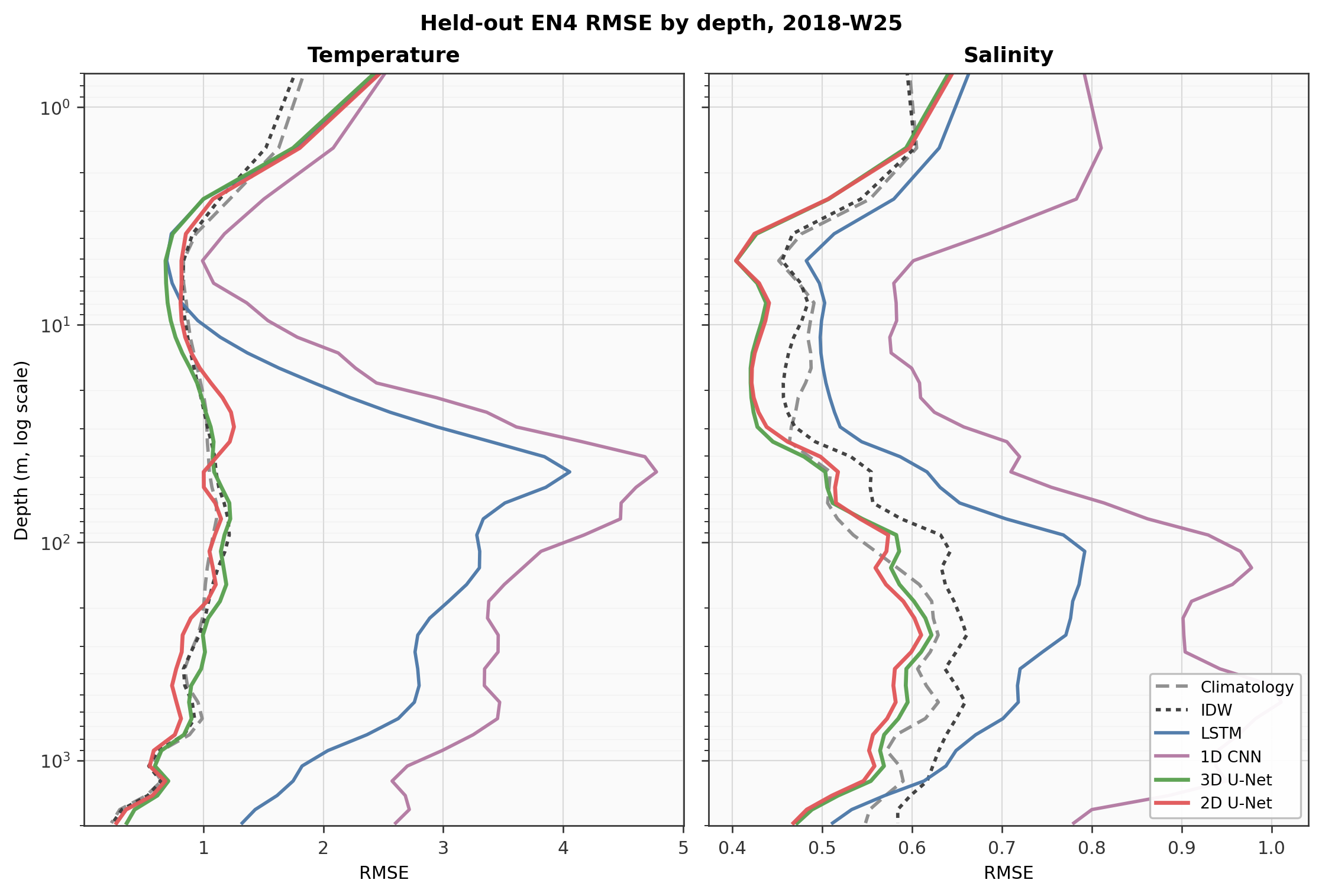}
    \caption{\textbf{RMSE per depth against held-out EN4 samples.} T and S reconstruction error by depth level, averaged over the global 2018 week-25 validation set.}
    \label{fig:res_graph}
\end{figure}

% TODO: Fill in values once baselines are trained.

\Cref{tab:recon_results} summarizes the reconstruction quality on held-out EN4 profiles, averaged over depth levels. The results show that predicting the climatology and IDW remain strong temperature baselines, especially for RMSE and MAE, indicating that much of the weekly thermal structure is already captured by the seasonal background and nearby profiles. 
The learned U-Net baselines are nevertheless competitive for temperature and improve the explained variance.
The benefit of spatial context is clearer for salinity, where both U-Net variants outperform the point-wise LSTM and 1D CNN, and where the 2D U-Net achieves the best scores overall. 
This suggests that salinity reconstruction benefits more from horizontal spatial context (such as water-mass structure and fronts) to reconstruct coherent patch fields.
In contrast, column-wise models are limited to local vertical priors. 
\autoref{fig:res_graph} shows that the U-Net baselines remain more stable across upper and intermediate ocean depth levels, while point-wise baselines exhibit larger errors which increase at deeper levels.

% \subsection{Surface Forecasting}
% \label{sec:forecast_results}

% \begin{table}[t]
% \centering
% \caption{\textbf{SST forecasting results} (validation 2018).
% Patch-level RMSE (K) at 1-, 2-, and 4-week lead times.
% \label{tab:forecast_results}}
% \small
% \begin{tabular}{@{}lccc@{}}
% \toprule
% & \multicolumn{3}{c}{\textbf{RMSE (K)}} \\
% \cmidrule(lr){2-4}
% \textbf{Method} & +1\,wk & +2\,wk & +4\,wk \\
% \midrule
% Persistence & -- & -- & -- \\
% Climatology & -- & -- & -- \\
% ConvLSTM & -- & -- & -- \\
% U-Net (stacked) & -- & -- & -- \\
% \bottomrule
% \end{tabular}
% \end{table}

% TODO: Fill in values once baselines are trained.

% \Cref{tab:forecast_results} shows SST forecasting performance.
% Persistence is a strong baseline at short lead times but degrades at longer horizons, where learned methods that capture spatiotemporal patterns show increasing advantage.
% These results confirm that \datasetname's weekly temporal structure contains learnable dynamics beyond what static baselines capture.

%% file: sec/6_conclusion.tex
\section{Conclusion}
\label{sec:conclusion}

We have introduced \datasetname, an open, global, AI-ready dataset of paired satellite surface and \emph{in situ} subsurface ocean observations spanning several decades.
By co-locating SST, SSS, and ADT with Argo profiles and matched GLORYS12~\cite{glorys12} reanalysis data at \SI{0.1}{\degree} spatial resolution and weekly \added{intervals} over 2000--2024 (9.5\,M profiles, 1283 weekly dates), \datasetname provides a unified resource that has the potential to support a range of ocean-related ML tasks.
We have demonstrated that the task of subsurface ocean state reconstruction, one task that is enabled by \datasetname, provides a challenging ML setting where predicting the climatology often performs better than simple ML baselines.
% We demonstrated this versatility through two complementary benchmarks---subsurface reconstruction from surface observations and sea surface temperature forecasting---and established baseline results for both.
% Our comprehensive survey of existing datasets (\cref{sec:related_work}) shows that no prior open resource combines raw paired observations, 0.1$^\circ$ resolution, weekly frequency, both temperature and salinity, global coverage, and dual observational/reanalysis ground truth.
% Beyond the two tasks benchmarked here, \datasetname could support research into regional and global spatiotemporal dynamics, mixed layer depth estimation, water mass classification, ocean heat content monitoring, and data assimilation---any application that benefits from co-located surface and subsurface observations at high resolution.
We hope that \datasetname can further spur research at the intersection of ML and oceanography.

\paragraph{Limitations.}
\datasetname inherits the sampling biases of the Argo network: coverage is sparser in marginal seas, near coasts, under ice, and below 2{,}000\,m. 
% The uint8 quantization introduces small encoding errors, negligible relative to measurement uncertainty. While t
While the reliance on L4 products, the depth-wise interpolation, and weekly aggregation make the dataset much easier to use and create a shared baseline setup for future experiments, this type of aggregation is not necessarily optimal for every user.

% \paragraph{Future work.}
% Planned extensions include adding biogeochemical variables from BGC-Argo, incorporating additional satellite products (ocean color, surface currents), extending temporal coverage with Deep Argo and legacy hydrographic data, and defining further evaluation tracks for downstream applications.

\paragraph{Availability.}
The dataset (\(\qty{\sim120}{\gibi\byte}\)), code, and reproducible loading examples are publicly available on {Hugging Face} under a \texttt{CC BY 4.0} license: \url{https://huggingface.co/datasets/ESA-philab/OceanDepths}.
%and GitHub. \footnote{\href{{https://anonymous.4open.science/r/OceanVariableReconstruction}}{\url{https://anonymous.4open.science/r/OceanVariableReconstruction}}}
% https://huggingface.co/datasets/simon-donike/OceanVariableReconstruction
%https://anonymous.4open.science/r/OceanVariableReconstruction/
% All data export and processing scripts, along with \texttt{pytorch} DataLoader code, 
% % can be found in an anonymized repository and 
% which are made publicly available \added{under a permissive \texttt{CC BY 2.0} license}: \url{https://huggingface.co/datasets/ESA-philab/OceanDepths}.

\paragraph{Acknowledgements.} \added{We thank the Met Office Hadley Centre for EN4.2.2~\cite{en4} and OSTIA, the Consiglio
Nazionale delle Ricerche for MULTIOBS, CLS for DUACS, and Mercator Ocean International for
GLORYS12V1~\cite{glorys12}. This study used E.U. Copernicus Marine Service Information.
The Argo observations were collected and made freely available by the International Argo
Program and its contributing national programs; Argo is part of the Global Ocean Observing
System (\href{https://doi.org/10.17882/42182}{doi:10.17882/42182}).
LLMs (Claude Code) were used for writing and coding assistance. We remain fully responsible for the conceptualization and execution of the research described in this manuscript, and we confirm that it describes our work completely and accurately. We carefully verified the content and all references.
}

%% file: sec/7_appendix.tex
\section{Appendix: Extra figures}

\begin{figure}
    \centering
    \includegraphics[width=\linewidth]{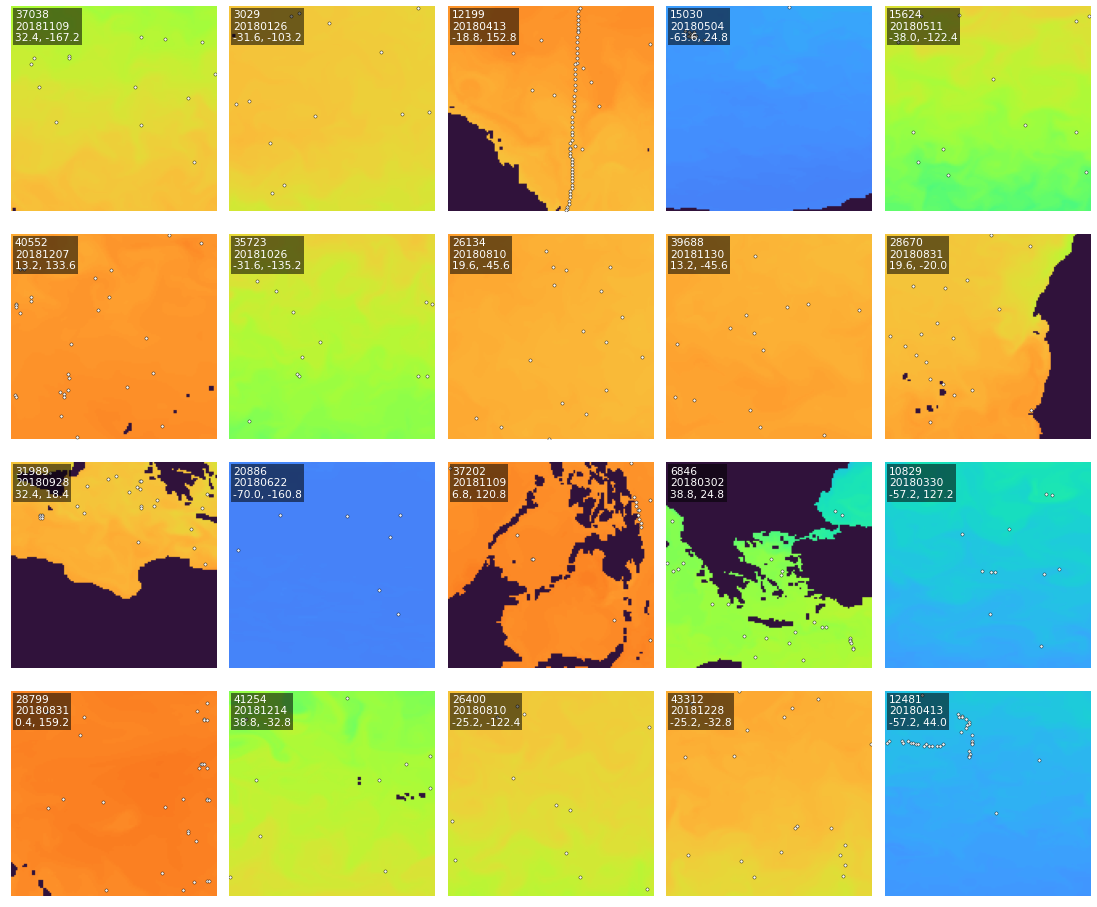}
    \caption{\textbf{Random selection of surface-level training patches} from the dataset, illustrating the diversity of ocean regions, SST gradients, and coastline configurations encountered during training.}
    \label{fig:random_samples}
\end{figure}

\iffalse
%this should be percentage comparison, how many do we keep, to be meaningful. but thats another mental detout to describe in the text.
    \begin{figure}
        \centering
        \includegraphics[width=\linewidth]{figures/glorys_target_alignment_within_cutoff_fraction.png}
        \caption{\textbf{Fraction of profiles accepted at each GLORYS depth level} under the depth-adaptive acceptance criterion. Near-surface levels (indices 0--10) retain $>$97\% of profiles, while acceptance drops below 10\% for the deepest levels (indices $>$38), reflecting the decreasing density of Argo observations with depth.}
        \label{fig:acceptance_fraction}
    \end{figure}
\fi

\begin{figure}
    \centering
    \includegraphics[width=\linewidth]{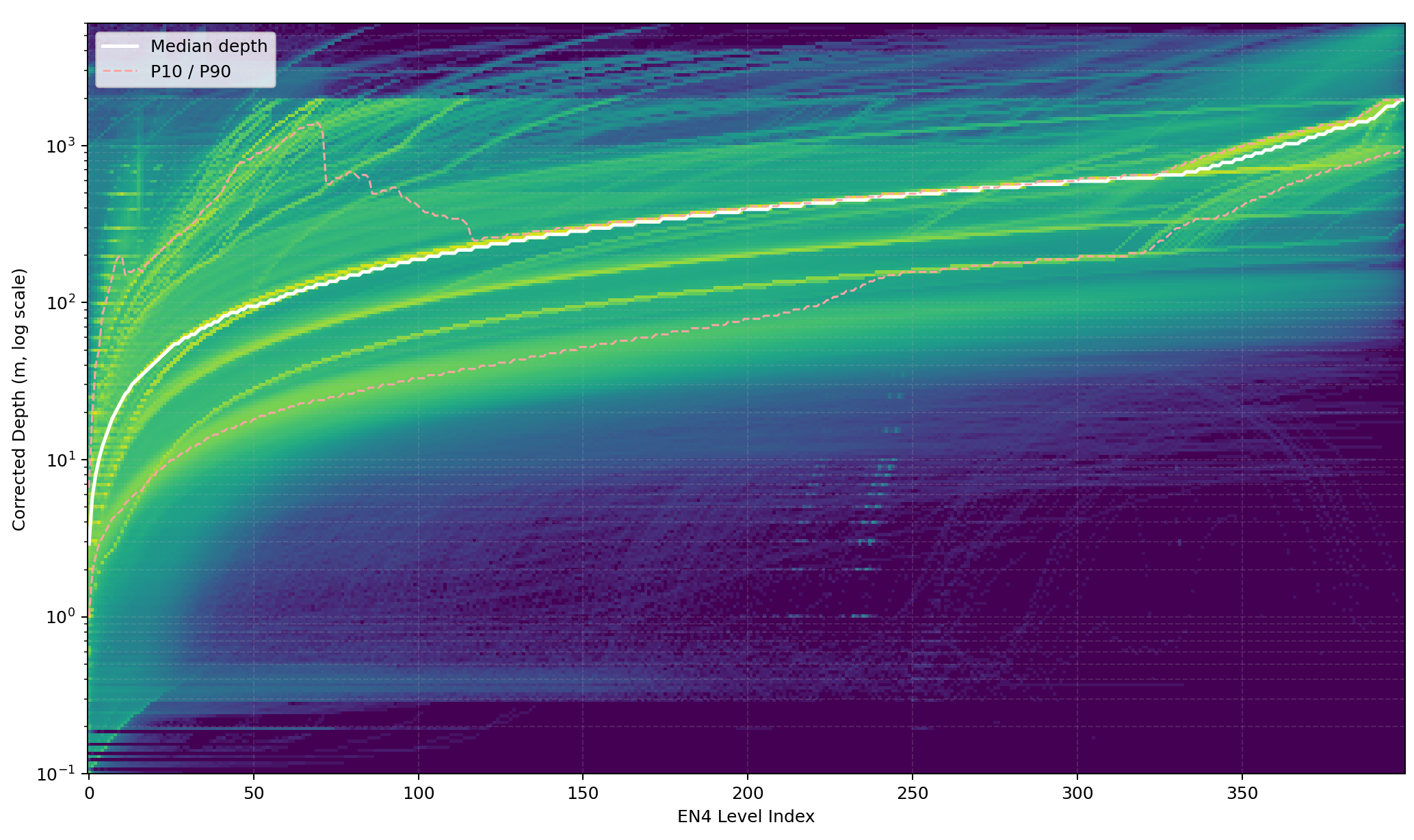}
    \caption{\textbf{EN4 depth index compared to depth in meters.}
    Heatmap of corrected depth (m, log scale) vs.\ EN4 level index, with median (white) and P10/P90 (pink dotted lines) curves. The irregular, profile-dependent depth sampling motivates the interpolation onto the fixed 50-level GLORYS12 coordinate.}
    \label{fig:depth_distribution}
\end{figure}